\documentclass[submission,copyright,creativecommons]{eptcs}
\providecommand{\event}{ICLP DC 2023} 

\usepackage{iftex}

\ifpdf
  \usepackage{underscore}         
  \usepackage[T1]{fontenc}        
\else
  \usepackage{breakurl}           
\fi

\usepackage{amsmath}
\usepackage{amsthm}
\usepackage{amssymb}
\usepackage{tikz}
\usetikzlibrary{positioning}
\usepackage{caption}
\usepackage{subcaption}
\usepackage[utf8]{inputenc}
\usepackage{titlesec}
\titlespacing*{\paragraph}{0pt}{1.25ex plus 1ex minus .2ex}{1em}

\newtheorem{example}{Example}

\title{Explainable Answer-set Programming}
\author{Tobias Geibinger
\institute{Knowledge-based Systems Group, 
Institute for Logic and Computation, 
TU Wien, Austria\\}
\email{tobias.geibinger@tuwien.ac.at}}
\def\titlerunning{Explainable Answer-set Programming}
\def\authorrunning{Tobias Geibinger}

\begin{document}

	

	

\maketitle

\begin{abstract}
  The interest in explainability in artificial intelligence (AI) is growing vastly due to the near ubiquitous state of AI in our lives and the increasing complexity of AI systems. Answer-set Programming (ASP) is used in many areas, among them are industrial optimisation, knowledge management or life sciences, and thus of great interest in the context of explainability. To ensure the successful application of ASP as a problem-solving paradigm in the future, it is thus crucial to investigate explanations for ASP solutions. Such an explanation generally tries to give an answer to the question of why something is, respectively is not, part of the decision produced or solution to the formulated problem. Although several explanation approaches for ASP exist, almost all of them lack support for certain language features that are used in practice. Most notably, this encompasses the various ASP extensions that have been developed in the recent years to enable reasoning over theories, external computations, or neural networks. This project aims to fill some of these gaps and contribute to the state of the art in explainable ASP. We tackle this by extending the language support of existing approaches but also by the development of novel explanation formalisms, like contrastive explanations. 
\end{abstract}

\section{Introduction and Problem Definition}
The topic of ``explainability'' in \emph{artificial intelligence} (AI) has become increasingly prominent over the recent years and regulation is being discussed.

Nowadays, there is a multitude of AI systems in use for decision-making and problem-solving. In this project we concern ourselves with \emph{Answer-set Programming} (ASP)~\cite{eiter09} which is a popular declarative problem-solving paradigm used in many domains~\cite{erdem16,falkner18}. ASP allows for the encoding of problems in a succinct way. Such a problem encoding usually consists of a set of \emph{rules} representing the underlying constraints of the problem and sets of \emph{facts} describing concrete scenarios. The solutions are given in terms of \emph{answer-sets}, which capture the evaluation of the rules. 

Extensions of ASP have been considered to model and solve problems in practice for a variety of reasons. Sometimes language extensions ease the modelling of the problem and make the representation more concise. However, ASP extensions can also enable new forms of reasoning and problem-solving, like for example, over custom theories~\cite{gebser16}, external computations~\cite{eiter16,gebser14}, or neural networks~\cite{yang20}.

The same reasons that make ASP a popular problem-solving paradigm, also make it attractive in the context of explainable AI. ASP is a symbolic and rule-based approach and thus has good prerequisites for humanly readable and intuitive explanations. In ASP, they usually amount to justifications as to why certain elements are (or are not) contained in an answer-set or why there is no answer-set at all. 

While there is a rich body of work on explainability for standard ASP~\cite{fandinno19}, the same is not the case for its numerous extensions. In fact, most current explanation approaches do not directly support basic language features like variables or disjunction. Furthermore, the notion of \emph{contrastive explanations} has recently been brought to attention to the AI community by Miller~\cite{miller2019}, who argues in favour of them due to their established history in the psycho- and sociological communities. 

The goal of this project is thus to close those gaps by focusing on (contrastive) notions of explainability for ASP extensions.

\section{Background}\label{sec:background}

	
	\paragraph{Answer-set Programming}
We consider disjunctive \emph{Answer-set Programming} (ASP)~\cite{eiter09}. An ASP program is a (finite) set of (disjunctive) rules of the form $a_1 \lor \dots \lor a_n \leftarrow b_1, \dots , b_k, \mathit{not} \ b_{k+1}, \dots , \mathit{not} \ b_m $,
where all $a_1,\dots a_n$ and $b_1,\dots ,b_m$ are function-free first-order atoms.
The head 
is the set of atoms  $a_1,\dots a_n$ before the implication symbol $\leftarrow$, and the body are the atoms and negated atoms $b_1,\dots ,b_m$.
The intuitive meaning of a rule is that if all atoms 
${b}_{1}, \ldots, {b}_{k}$
can be derived, and there is no evidence for any of the atoms 
$b_{k+1}, \ldots, b_{m}$
(i.e., the rule fires), then at least one of 
$a_{1}, \ldots, a_{n}$
has to be derived.
A rule with an empty body (i.e., $m=n=0$) is called a \emph{fact},
with $\leftarrow$ usually omitted and a rule with empty head (i.e., $n=0$) is a \emph{constraint}. Furthermore, whenever the head of a rule is a singleton, we call the rule \emph{normal}.
The semantics is defined as usual in terms of particular \emph{(Herbrand) models} of the \emph{grounding} of the program and an \emph{answer-set} $I$ is a $\subseteq$-minimal model of the Gelfond-Lifschitz reduct $P^I = \{ a_1 \lor \dots \lor a_n \leftarrow b_1, \dots , b_k \mid r 
\in P, \ \{ b_{k+1},\ldots b_{m} \} \cap I = \emptyset \}$.

	Several ASP language extensions exist, some are mainly syntactic sugar enabling more concise representation of problems, like \emph{aggregates}~\cite{faber04} or \emph{choice rules}~\cite{gebser12}. However, others either extend the semantics of ASP or generalise them to full propositional theories as \emph{equilibrium logic}~\cite{pearce06} does. Extensions of the semantics include reasoning over linear constraints~\cite{banbara17} or arbitrary theories~\cite{gebser14}, but they also facilitate neuro-symbolic computation, like for example, \emph{NeurASP}~\cite{yang20} or $\mathit{LP}^\mathit{MLN}$ programs~\cite{lee16}.
\vspace{-1mm}
\paragraph{Explanations for ASP}
	
Several approaches for explaining consistent ASP programs exist. The goal is usually to give a justification as to why an atom is contained respectively not contained in an answer-set. 
\emph{Off-line justifications}~\cite{pontelli09,trieu2021,trieu2022} are labelled directed graphs providing reasons for why an atom is (or is not) contained in a given answer-set. Intuitively, this can be seen as a derivation of the truth value of the respective atom using the rules of the program. 
\emph{Causal justifications}~\cite{fandinno16} provide explanations for literals in a given answer-set by means of an alternative semantics for logic programs with disjunction. 	
The basis for causal justifications are \emph{causal terms} which are algebraic terms representing joint causation through multiplication and alternative causes through addition. 
\emph{Why-not provenance}~\cite{damasio13} provides justifications for the truth values of atoms w.r.t answer-set semantics for normal logic programs. In particular, it explains literals in a general fashion for a given program without the need of a specific answer-set.
\emph{Witnesses}~\cite{wang2022} give reason as to why an atom is contained in a particular answer-set by giving a resolution proof. 
Similary, the system \texttt{xclingo}~\cite{cabalar20} prints out a rule derivation for the atom to be explained. 
Explanations for inconsistent programs are usually generated to help with debugging programs and make them consistent. 
For example, the systems \texttt{spock}~\cite{gebser08} and \texttt{Ouroboros}~\cite{oetsch10} explain why an interpretation is not an answer-set via program transformations which encode the given ASP program into a meta-program. Answer-sets of this program contain an interpretation which is not an answer-set and the reasons why. 
Saribatur et al.~\cite{saribatur21} employed \texttt{Ouroboros} in their \emph{abstraction} method and used it to generate explanations for inconsistent instances in certain problem domains.
Further debugging approaches are DWASP~\cite{dodaro15} and \texttt{stepping}~\cite{oetsch18}. DWASP was introduced for the ASP solver WASP~\cite{alviano13} and works by amending \emph{debug atoms} to programs. The inconsistency of the resulting \emph{debugging program} can then be explained in terms of sets of debug atoms that cannot jointly hold. 
In difference, \texttt{stepping} takes lets the user perform the solving process by applying one applicable rule after the other.

\vspace{-1mm}
\paragraph{Contrastive Explanation}
To answer questions like ``Why P and rather than Q?'' is the basis of contrastive explanations \cite{lipton1990}.
It has been argued that such explanations are intuitive for humans to understand and to produce and also that standard why questions contain a hidden \emph{contrast case}, e.g., ``Why $P$?'' represents ``Why $P$ rather than not $P$?''~\cite{miller2019}.
Lipton~\cite{lipton1990} defines an answer to such a question as the \emph{difference condition}, stating that the answer contains a cause for $P$ that is missing for not $Q$. 
\begin{example}\label{ex:bugs}
	Consider an algorithm that classifies bugs and suppose the algorithm has classified a particular bug as a beetle. A potential non-contrastive explanation for this classification is to present the values of the features, i.e., the bug is a beetle because it has 8 legs, 2 eyes and 2 wings. If we interpret the question ``Why is the bug a beetle?'' as ``Why is the bug a beetle instead of another bug?'', then an adequate explanation is to highlight that the bug has 2 eyes, if it had 5 eyes, then it would be a fly.
\end{example}

\section{Research Goals}\label{sec:goals}
	
	\paragraph{Explaining ASP Extensions and Advanced Features}
	One of our main research directions will be the study of explainability in the context of the ASP extensions, as the mentioned existing explanation approaches often do not support them. 
	Besides existing approaches, we intend to investigate explanations based on \emph{abduction}, i.e. the process of determining which hypothesis need to be added to a theory to explain given observations. This reasoning method has been studied for logic programs~\cite{eiter97} and -- depending on the use-case -- can be also be employed as an explanation approach. This seems interesting in the context of contrastive explanations~\cite{lipton1990}, which can be expressed through abductive reasoning.~\cite{silva23}

	Furthermore, it is worthwhile to investigate the different ways of how the ``inner workings'' of extensions can be considered in the explanations. We intend to consider the following settings. 
	In a black-box setting the explanation approach does not take into account how the ASP extension determines the truth values of its atoms. For HEX programs, this would mean that external atoms are simply considered true or false depending on the context and their truth values. The direct opposite to the previously mentioned approach would be the white-box setting, where the semantics of the ASP extension is fully known and considered in the explanation. In the case of NeurASP, the explanation would need to incorporate information about the neural network. The middle way between the black-box and the white-box approach would be a gray-box setting, where we do not have full access to the inner workings of the ASP extension but we do have some information. For external computations or neural atoms, the explanation approach could take, for example, certain input/output relationships into account. 

	\paragraph{Explaining Instead of Debugging Inconsistency}
	The explanation approaches for inconsistent programs mentioned in Section~\ref{sec:background} are all intended to facilitate debugging of answer-set programs. While those explanations focused on debugging are of course all valid, they may not be useful in all contexts, for example when the explanation is needed by an end-user and not an ASP engineer. Our aim is thus to investigate explanation approaches for inconsistency that are more akin to the ones for consistent programs. Related work in this direction was done by Dam{\'{a}}sio et al.~\cite{damasio15} who extended the debugging system \texttt{spock} with providing justifications for atoms in the case that the program is actually consistent.

	\paragraph{Explainability and Equilibrium Logic}
	Certainly, a formal proof can be seen as an explanation as to why some proposition has to follow from the given premises. In equilibrium logic~\cite{pearce06} this is not different.
	While a proof system for equilibrium logic based on tableaux methods exists~\cite{pearce00}, there are two main issues that make it problematic for usage in explainability. The first is that the tableaux approach contains multiple steps and it is difficult to obtain humanly readable explanations from the tableaux. The other is that the axiomatized entailment relation is similar to what is also called \emph{skeptical inference}, i.e. the inference holds if the proposition is contained in each equilibrium model (answer-set). 
	However, as we have seen in Section~\ref{sec:background}, explanation approaches generally concern themselves with the question of why something is true in \emph{some} answer-set and this case is arguably also more important in practice. The corresponding entailment relation is often called \emph{brave} or \emph{credulous inference}. 
	We thus intend to develop proof systems for this type of inference that are more akin to the more understandable sequent calculi, similar to what was done for related nonmonotonic formalisms~\cite{bonatti96}. 
	Furthermore, abduction has also been investigated for equilibrium logic~\cite{pearce01} and we intend to investigate abductive explanations.


	\paragraph{Towards Practical Algorithms}
	A natural step in our study of explainability for ASP is to produce implementations for the methods we introduce. An approach that is often employed in generating inconsistent explanations, like for \texttt{spock}~\cite{gebser08} and \texttt{Ouroboros}~\cite{oetsch10}, is to encode this in ASP itself. 
	We also aim to provide implementations of the approaches we introduce in certain problem domains, like scheduling or planning, where ASP has been shown to produce good solutions and benchmark data is available. Specific questions may arise in those areas and whenever possible we aim to investigate tailored enhancements for said domains. 
	Any developed explanation system should also be interactive, as it has been argued that the act of explanation is not static and more of a dialogue~\cite{miller2019}.
	Hence, we aim to build systems, where the user can guide the explanation process.

\section{Research Status \& Outlook}
The project was officially started in the summer of 2022 and is expected to last for 3 years. 
Currently, we are focused on two research directions. The first is investigating formal notions of justification, akin to off-line justifications~\cite{pontelli09} and witnesses~\cite{wang2022}, for disjunctive programs comprised of Abstract Constraint Atoms~\cite{marek2004}. Programs defined over such atoms neatly generalise several language extensions. 
\begin{example}
	Consider the aggregate $\mathtt{\#sum\{ 2:a, 1:b, 1:c \} > 1 }$. Suppose we have a model $I_1 = \{ a, b, c \}$. 
	Then, an intuitive justification as to why $I_1$ is a model of the aggregate, is that it is because $a \in I_1$. Of course, it is also valid to say that $I_1$ is a model because $b,c \in I_1$.
Consider $I_2 = \{b\}$. Since $I_2$ is not a model of the aggregate, we would like to know why. The justification in this case is that neither $a$ nor $c$ are satisfied by $I_2$, which, since $b$ cannot be false in $I_2$, would be a requirement.
\end{example}
As the example shows, this approach is purely based on semantics and does not take rule application into account. Therefore, we are also working on more syntactic notions that can be used together with the above concept is to use them both in an interactive explanation system that supports choice, aggregates and disjunctive rules. A conference paper on this topic has recently been accepted~\cite{eiter23a}.

The other direction we are pursuing is that of contrastive explanations with and for ASP. 
\begin{example}
	Consider the bug classification from Example~\ref{ex:bugs} and suppose it is expressed by an ASP program $P$. So, given the instance $F=\{ \mathit{legs}(6),\mathit{eyes}(2), \mathit{wings}(2) \}$, we obtain the answer-set $I = \{ \mathit{class}(\mathit{beetle}) \}$. The question ``Why is the bug a beetle instead of another bug?'' can then be formulated as the problem of finding some $F'$ such that $I$ is not the answer-set of $P\cup F'$ and $F'$ differs minimally from $F$. In this case, $F' = \{  \mathit{legs}(6),\mathit{eyes}(5), \mathit{wings}(2) \}$, $F\setminus F' = \{\mathit{eyes}(2)\}$ and $F'\setminus F = \{\mathit{eyes}(5)\}$.
\end{example}
In the example we said that the contrastive set of facts should be minimally different. A natural choice would be minimal symmetric difference, but in general this difference is problem-dependent. In the example the rules remained fixed, which is not always suited. For example, Bogatarkan et al.~\cite{bogatarkan2020} use similar explanations for their path finding system, but they focus on relaxing constraints. On this subject, a paper, focused on a specific application, has just been accepted~\cite{eiter23b} and another is under review.

Besides the numerous research goals we have not tackled yet, there are some bigger issues with the overall topic of explainability. When Miller~\cite{miller2019} surveyed the work on explainability, he noticed that formal notions of explainability often differ from what is accepted as explanation in the social sciences. This includes explanation approaches that are purely static mathematical objects and Miller thus argues for interactive approaches that may or may not utilise contrastiveness.

Regarding expected achievements, we plan to investigate the research topics presented in Section~\ref{sec:goals} and implement them in an explanation system to offer interactive explanations. 

\bibliographystyle{eptcs}
\bibliography{references}

\end{document}